\documentclass[10pt,twocolumn,letterpaper]{article}

\usepackage{iccv}
\usepackage{times}
\usepackage{epsfig}
\usepackage{graphicx}
\usepackage{amsmath}
\usepackage{amssymb}
\usepackage{multirow}
\usepackage{booktabs} 
\usepackage{subcaption}
\usepackage{balance}

\newcommand\blfootnote[1]{%
  \begingroup
  \renewcommand\thefootnote{}\footnote{#1}%
  \addtocounter{footnote}{-1}%
  \endgroup
}

\usepackage[pagebackref=true,breaklinks=true,letterpaper=true,colorlinks,bookmarks=false]{hyperref}

\newcommand{\shape}{\mbox{\boldmath $\beta$}}
\newcommand{\pose}{\mbox{\boldmath $\theta$}}
\newcommand{\vidt}{\mbox{\boldmath $\Theta$}}
\newcommand{\cam}{\mbox{\boldmath $\pi$}}

\iccvfinalcopy 

\def\iccvPaperID{****} 

\ificcvfinal\fi
\begin{document}

\title{TexturePose: Supervising Human Mesh Estimation with Texture Consistency}

\author{Georgios Pavlakos\textsuperscript{*}, Nikos Kolotouros\textsuperscript{*}, Kostas Daniilidis\\[0ex]
University of Pennsylvania
}

\maketitle
\ificcvfinal\pagestyle{empty}\fi

\begin{abstract}
This work addresses the problem of model-based human pose estimation. Recent approaches have made significant progress towards regressing the parameters of parametric human body models directly from images. Because of the absence of images with 3D shape ground truth, relevant approaches rely on 2D annotations or sophisticated architecture designs. In this work, we advocate that there are more cues we can leverage, which are available for free in natural images, i.e., without getting more annotations, or modifying the network architecture. We propose a natural form of supervision, that capitalizes on the appearance constancy of a person among different frames (or viewpoints). This seemingly insignificant and often overlooked cue goes a long way for model-based pose estimation. The parametric model we employ allows us to compute a texture map for each frame. Assuming that the texture of the person does not change dramatically between frames, we can apply a novel texture consistency loss, which enforces that each point in the texture map has the same texture value across all frames. Since the texture is transferred in this common texture map space, no camera motion computation is necessary, or even an assumption of smoothness among frames. This makes our proposed supervision applicable in a variety of settings, ranging from monocular video, to multi-view images. We benchmark our approach against strong baselines that require the same or even more annotations that we do and we consistently outperform them. Simultaneously, we achieve state-of-the-art results among model-based pose estimation approaches in different benchmarks. The project website with videos, results, and code can be found at {\footnotesize \url{https://seas.upenn.edu/~pavlakos/projects/texturepose}}.  
\blfootnote{$^*$ equal contribution}
\end{abstract}

\section{Introduction}
\begin{figure}[!t]
	\centering
	\includegraphics[scale=0.3,trim={60mm 30mm 60mm 20mm},clip]{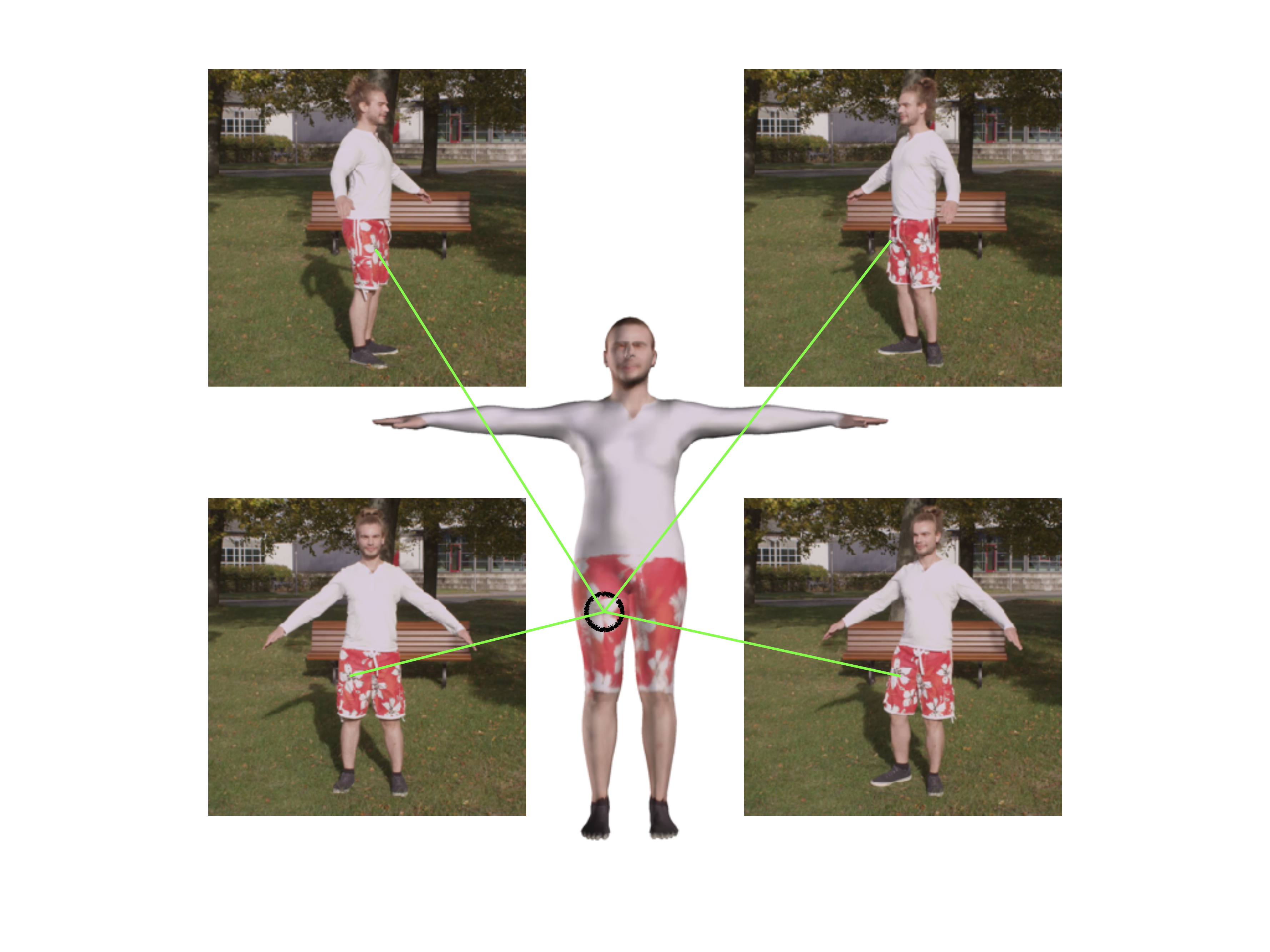}
	\caption{For a short video, or multi-view images of a person, a specific patch on the body surface has constant texture. This consistency can be formulated as an auxiliary loss in the training of a network for model-based pose estimation, and allows us to leverage information directly from raw pixels of natural images. Images and texture come from the People-Snapshot dataset~\cite{alldieck2018video}.}
\label{fig:firstpage}
\end{figure}

\begin{figure*}[!t]
	\centering
	\includegraphics[scale=0.45]{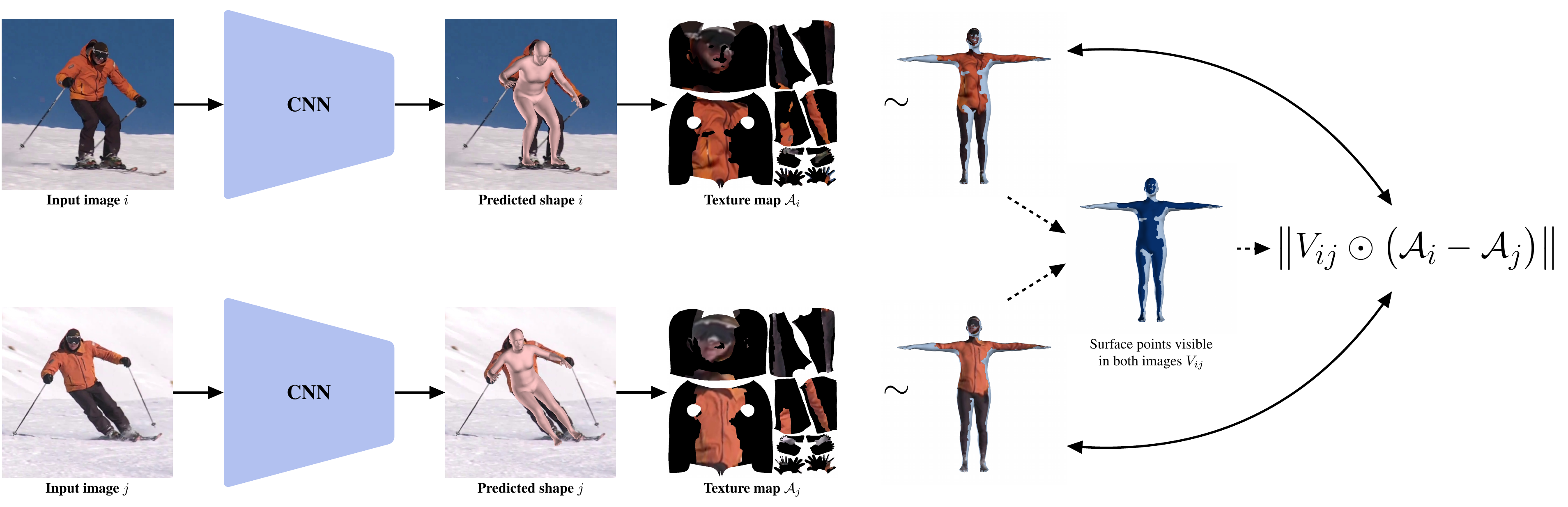}
	\caption{Overview of the proposed texture consistency supervision. Here, for simplicity, the input during training consists of two images $i,j$ of the same person. The main assumption is that the appearance of the person does not change dramatically across the input images, (i.e., the frames come from a monocular video as in Figure~\ref{fig:firstpage}, or from time-synchronized multi-view cameras). We apply our deep network on both images and estimate the shape of the person. Subsequently, we project the predicted shape on the image, and after inferring visibility for each point on the surface, we build the texture maps $\mathcal{A}_i$ and $\mathcal{A}_j$. The crucial observation, that the appearance of the person remains constant, translates to a texture consistency loss, forcing the two texture maps to be equal for all surface points $V_{ij}$ that are visible in both images. This loss acts as supervision for the network and complements other weak losses that are typically
used in the training.}
\label{fig:teaser}
\end{figure*}

In recent years, the area of human pose estimation has experienced significant successes for tasks with an increasing level of difficulty; 2D joint detection~\cite{newell2016stacked,wei2016convolutional}, dense correspondence estimation~\cite{alp2018densepose} or even 3D skeleton reconstruction~\cite{martinez2017simple,sun2018integral}. Typically, as we ascend the pyramid of human understanding, we target more and more challenging tasks. As expected, the emergence of sophisticated parametric models of the human body, like SCAPE~\cite{anguelov2005scape}, \mbox{SMPL(-X)}~\cite{loper2015smpl,pavlakos2019expressive,romero2017embodied}, and Adam~\cite{joo2018total,xiang2019monocular}, has really paved the way for full 3D pose and shape estimation from image data. And while this step has been well explored for video or multi-view data~\cite{huang2017towards,joo2018total}, the ultimate goal is to reach the same level of analysis from a single image.

Traditional optimization-based approaches, e.g.,~\cite{bogo2016keep,guan2009estimating,lassner2017unite}, have performed very reliably for model-based pose estimation. However, more recently, the interest has moved  towards data-driven approaches regressing the parameters of the human body model, directly from images. Considering the lack of images with 3D shape ground truth for training, the main challenge is to identify reliable sources of supervision. Proposed methods~\cite{kanazawa2018end,omran2018neural,pavlakos2018learning,tan2017indirect,tung2017self,zanfir2018deep} have focused on leveraging all the available sources of 2D annotations like 2D keypoints, silhouettes, or semantic parts. Simultaneously, external sources of 3D data (e.g., MoCap and body scans) can also be useful, by applying learned priors~\cite{kanazawa2018end}, or decomposing the task in different architectural components~\cite{omran2018neural,pavlakos2018learning,zanfir2018deep}. In this work, instead of focusing on the available 2D annotations, or the appropriate way to employ external 3D data, the questions we ask are different. Can natural images alone provide us a useful cue for this task? Is there a form of supervision we can leverage without further annotations? Here, we argue, and demonstrate, that the answer to these questions is positive.

We present TexturePose, a way to leverage complementary supervision directly from natural images (Figure~\ref{fig:teaser}). The main observation is that the appearance of a person does not change significantly over small periods of time (e.g., during a short video). Our insight is that this appearance constancy enforces strong constraints in the estimated pose of each frame, which naturally translates to a powerful supervision signal that is useful for cases of monocular video or multi-view images. A critical component is the incorporation of a parametric model of the human body, SMPL~\cite{loper2015smpl}, within our pipeline, allowing us to map the texture of the image to a generic texture map, which is independent of the shape and pose. Considering a network estimating the model parameters, during training, we generate the mesh and project it on the image. Through efficient computation, we are able to infer a (partial) texture map for each frame. Our novel supervision signal, based on texture consistency, enforces that the texture of each point of the texture map remains constant for all the frames of the same subject. This seemingly unimportant piece of information goes a long way and proves itself to be a crucial form of auxiliary supervision. We validate its importance in settings involving multiple views of the same subject, or monocular video with very weak annotations. In every case, we compare with approaches that have access to the same level of annotations (or potentially even more), and we consistently outperform them. Ultimately, this supervision allows us to outperform state-of-the-art	approaches for model-based pose estimation from a single image.

Our contributions can be summarized as follows:
\begin{itemize}
	\item We propose TexturePose, a novel approach to leverage complementary supervision from natural images through appearance constancy of each human across different frames.
	\item We demonstrate the effectiveness of our texture consistency supervision in cases of monocular video and multi-view capture, consistently outperforming approaches with access to the same or more annotations than we do.
	\item We achieve state-of-the-art results among model-based 3D human pose estimation approaches.
\end{itemize}

\section{Related work}
In this Section, we summarize the approaches that are more relevant to ours. 

\textbf{Model-based human pose estimation}:
Differently from skeleton-based 3D pose estimation, model-based human pose estimation involves a parametric model of the human body, e.g., SCAPE~\cite{anguelov2005scape} or SMPL~\cite{loper2015smpl}. The goal is to estimate the model parameters that give rise to a 3D shape which is consistent with image evidence. The initial works in this area~\cite{guan2009estimating,sigal2008combined} as well some more recent approaches~\cite{bogo2016keep, lassner2017unite, zanfir2018monocular} were mainly optimization-based. Recently, the trend has shifted to directly regressing the model parameters from a single image using deep networks~\cite{kanazawa2018end,omran2018neural,pavlakos2018learning,zanfir2018deep}. Given the lack of images with 3D shape ground truth, these approaches typically rely on 2D annotations, like 2D keypoints, silhouettes and semantic parts, as well as external 3D data. Although, we believe there is great merit into using the bulk of already annotated data, in this paper we aspire to get beyond this data and explore complementary forms of supervision which are available also in unlabeled or weakly labeled data.

\textbf{Multi-view pose estimation}:
Our goal in this work is not explicitly to estimate human pose from multiple views (in fact the work of Huang~\etal~\cite{huang2017towards} addressed this nicely in a model-based way). However, our approach is relevant to recent approaches leveraging multi-view consistency as a form of supervision to train deep networks. Pavlakos~\etal~\cite{pavlakos2017harvesting} estimate 3D poses combining reliable 2D pose estimates, and treats them as pseudo ground truth to train a network for 3D human pose. Simon~\etal~\cite{simon2017hand} propose a similar approach to improve a hand keypoint detector given multi-view data. Rhodin~\etal~\cite{rhodin2018learning} learn 3D pose estimation by enforcing the pose consistency in all views. On the other hand, follow-up work from Rhodin~\etal~\cite{rhodin2018unsupervised}, uses multiple views to learn a representation of 3D human pose in an unsupervised manner. In contrast to the above works, we believe that our approach offers much greater opportunities to leverage multi-view consistency. The incorporation of a parametric model allows us to go beyond body joint consistency, by leveraging shape and texture consistency. Simultaneously, instead of learning a new representation from multi-view data, we choose to maintain the SMPL representation, and only leverage the collective power of data to better regress the parameters of this representation.

\textbf{Supervision signals}:
While we have already discussed some aspects of the supervision typically employed for 3D human pose estimation, here we attempt to extend the discussion particularly to the varying levels of supervision used by different works. Full body pose and shape supervision is typically available only in synthetic images~\cite{varol2017learning}, or images with successful body fits~\cite{lassner2017unite}. Weak supervision provided by 2D annotations is typical, with different works employing 2D keypoints, silhouettes and semantic parts~\cite{kanazawa2018end,omran2018neural,pavlakos2018learning,tan2017indirect}. Non-parametric approaches typically use extra supervision from 2D keypoint annotation~\cite{habibie2019wild,sun2018integral,zhou2017towards}, while some recent works leverage ordinal depth relations of the joints~\cite{pavlakos2018ordinal,ronchi2018s}. Multi-view consistency is also well explored as discussed earlier~\cite{kocabas2019self,pavlakos2017harvesting,rhodin2018unsupervised,rhodin2018learning,simon2017hand}. In terms of pose priors, Zhou~\etal~\cite{zhou2017towards} use weak symmetry constraints, while Kanazawa~\etal~\cite{kanazawa2018end} incorporates a learning-based prior on pose and shape parameters using adversarial networks. In contrast to the above, instead of using additional annotations or exploiting external information, our goal is to leverage all the information that is available in natural images. This of course does not exclude the use of other supervision forms. In fact, we demonstrate that our approach can properly complement typical supervision signals (e.g., 2D keypoints, pose priors), and improve performance only by additionally enforcing texture consistency.

\textbf{Texture-based approaches}:
The idea of using texture to guide pose estimation goes back at least to the work of Sidenbladh~\etal~\cite{sidenbladh2000stochastic}, where texture consistency was used for tracking. More recently, Bogo~\etal~\cite{bogo2017dynamic} use high resolution texture information to improve registration alignments. Guo~\etal~\cite{guo2017real} also enforce photometric consistency to recover accurate human geometry over time. Alldieck~\etal~\cite{alldieck2019learning,alldieck2018detailed,alldieck2018video} focus on estimating the texture for human models. In the work of Kanazawa~\etal~\cite{kanazawa2018learning}, texture is employed to learn a parametric model of bird shapes. While we share similar intuitions with the above works, here we propose to use texture as a supervisory signal to guide and improve learning for 3D human pose and shape estimation.

Finally, to put our work in a greater context, the idea of appearance constancy is popular also beyond human pose estimation, e.g., in approaches for unsupervised learning of depth, ego-motion and optical flow~\cite{jason2016back,meister2018unflow,ranjan2018adversarial,zhou2017unsupervised}. A key difference is that while they estimate the structure of the world in a non-parametric form (depth map), we instead inject some domain knowledge (i.e., assuming a human pose estimation task) and we leverage a model, SMPL, that helps us explain the image observations. A similar motivation is shared with the work of Tung~\etal~\cite{tung2017self}. However, our approach is more flexible, since they require keypoints as input to their network, frames should be continuous to allow for motion extraction, while they eventually rely on a separate network for optical flow computation. Simultaneously, we present a more generic framework, which can be applied for monocular video or multi-view images alike.

\section{Technical approach}
In this Section, we start with a short introduction about the representation we use and the basic notation (Subsection~\ref{basics}). Then, we describe the regression architecture (Subsection~\ref{regression}). We continue with the formulation of texture consistency, and the corresponding loss (Subsection~\ref{texcons}). Next, we describe the additional losses we can incorporate when we process images from monocular or multi-view input (Subsection~\ref{beyond}). Finally, we provide an overview of the complete pipeline (Subsection~\ref{complete}), and discuss potential weaknesses of our approach (Subsection~\ref{shortcomings}).

\subsection{Representation} \label{basics}
\textbf{SMPL}: 
The SMPL model~\cite{loper2015smpl} is a parametric model of the human body. Given the input parameters for pose $\pose$, and shape $\shape$, the model defines a function $\mathcal{M}(\pose, \shape)$ which outputs the body mesh $M \in \mathbb{R}^{3 \times N}$, with $N = 6890$ vertices. The body joints $X$ are expressed as a linear combination of the mesh vertices, so using a pre-trained linear regressor $W$, we can map from the mesh to $k$ joints of interest $X \in \mathbb{R}^{3 \times k} = WM$.

\textbf{Texture map}:
The meshes produced by SMPL are deformations of an original template $T$. A corresponding UV map un-warps the template surface onto an image, $\mathcal{A}$, which is the texture map. Each pixel $t$ of this texture map is also called texel. By construction, the mapping between texels and mesh surface coordinates is fixed and independent of changes in 3D surface geometry.

\textbf{Camera}:
The camera we use follows the weak perspective camera model. The parameters of interest are denoted with $\cam$ and include the global orientation $R \in \mathbb{R}^{3 \times 3}$, scale $s \in \mathbb{R}$, and translation $t \in \mathbb{R}^2$. Given these parameters, the 2D projection $x$ of the 3D joints $X$ is expressed as:
\begin{equation} \label{eq:proj}
x = \cam(X) = s\Pi(RX) + t,
\end{equation}
where $\Pi$ stands for the orthographic projection.

\subsection{Regression model} \label{regression}
Our goal is to learn a predictor $f$, here realized by a deep network, that given a single image $I$, it maps it to the pose and shape parameters of the person on the image. More concretely, the output of the network consists of a set of parameters $\vidt = f(I)$, where $\vidt= [\pose, \shape, \cam]$. Here, $\pose$ and $\shape$ indicate the SMPL pose and shape parameters, and $\cam$ are the camera parameters. Our deep network follows the architecture of Kanazawa~\etal~\cite{kanazawa2018end}, with the exception of the output, which in our case regresses 3D rotations using the representation proposed by Zhou~\etal~\cite{zhou2018continuity}.

\subsection{TexturePose} \label{texcons}
Given $\pose$ and $\shape$ we can generate a mesh $M$ and the corresponding 3D joints $X$. The mesh can be projected to the image using the estimated camera parameters $\cam$. Through efficient computation~\cite{psbody}, we can infer the visibility for each point on the surface, and as a result, for every texel $t$ of the texture map $\mathcal{A}$. Let us denote with $v_t$ the inferred visibility of texel $t$ on the texture map $\mathcal{A}$. The collection of all visibility indices $v_t$ can be arranged in a binary mask $V$, the visibility mask. Considering each point $P_t$ on the mesh surface, we can estimate its image projection using the camera parameters, $p_t = \cam (P_t)$. For the visible points, we can estimate their texture via bilinear sampling from the image $I$, so $a_t = G(I; p_t)$, where $G$ is a bilinear sampling kernel. The collection of all values $a_t$, i.e., texture values for every texel $t$, constitute the texture map $\mathcal{A}$.

Let us now assume that we have access to two images $i, j$ of the same person. Using the procedure above, we can estimate the two texture maps $\mathcal{A}_i, \mathcal{A}_j$, along with the corresponding visibility masks $V_i, V_j$. Let us denote with $V_{ij} = V_i \odot V_j$ the mask of the surface points that are visible in both views. Then the texture consistency loss can be simply defined as:
\begin{equation}
L_{\text{texture cons}} =  ||V_{ij} \odot (\mathcal{A}_i - \mathcal{A}_j)||.
\end{equation}
This loss enforces that the texture should be the same for texels (or equivalently, points on the surface) that are visible in both images. Since visibility masks are used only to mask-out the texels that should not contribute to the loss, visibility computation does not have to be differentiable.

\subsection{Beyond texture} \label{beyond}

\begin{figure}[!t]
	\centering
	\includegraphics[scale=0.32]{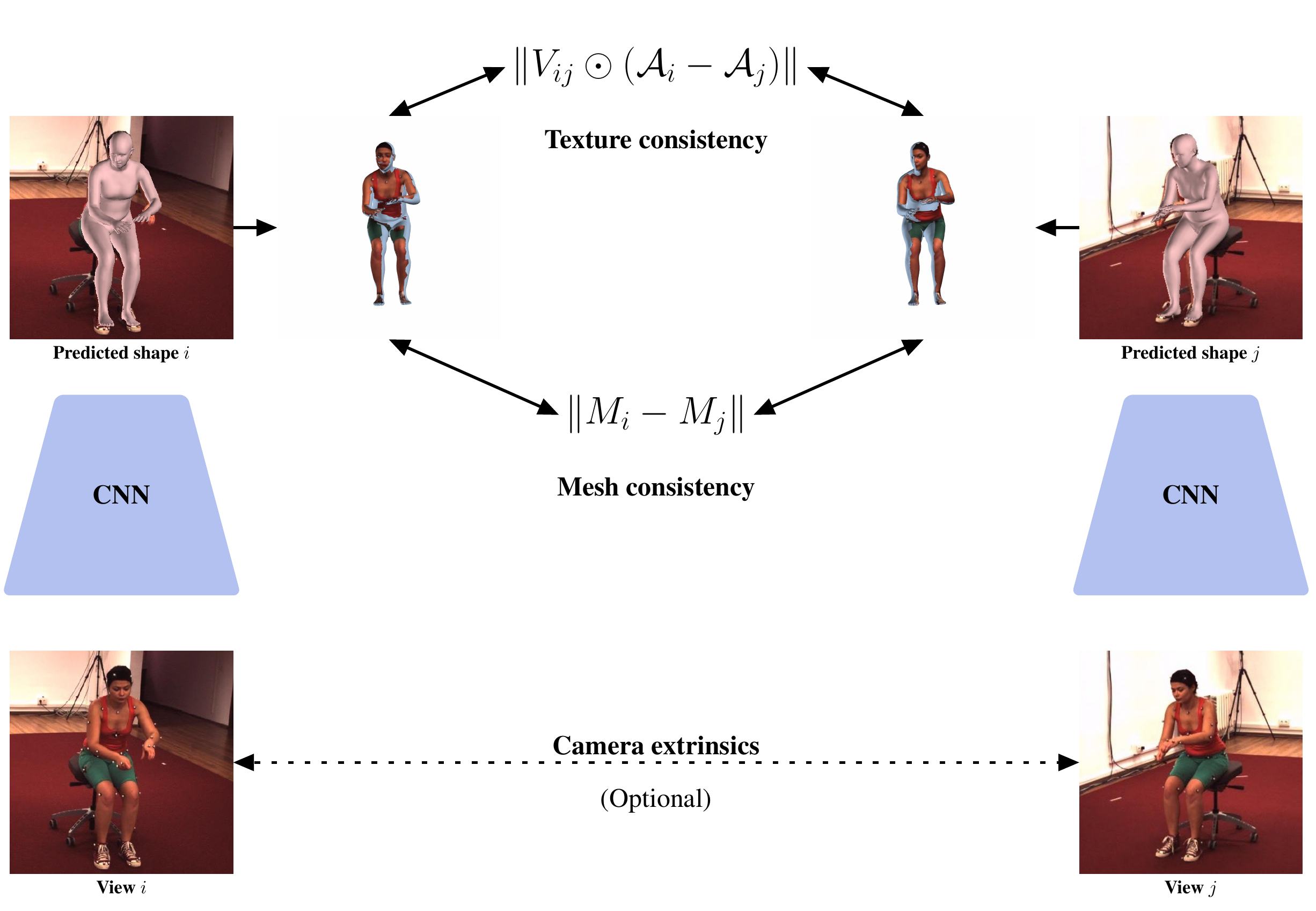}
	\caption{
With our formulation, training with images from a multi-camera system is similar to training with images from monocular video (Figure~\ref{fig:teaser}). The main additional consistency constraint is that the subject has the same 3D shape (same body mesh), which means that we can apply a per-vertex loss between the two mesh predictions. Before applying the predicted global orientation, the mesh predictions are in the same canonical orientation, so we can apply our loss directly on the mesh predictions. In case the extrinsics are provided, we can transform the second mesh to the frame of the first view, and then apply the same loss.}
\label{fig:mview}
\end{figure}

\textbf{Monocular}: 
In the monocular case, the texture consistency loss is applied between pairs of frames for the same subject. Beyond the texture consistency, we can also enforce that the shape parameters of the subject remain the same for all pairs of frames. This shape consistency can be enforced with the following loss function:
\begin{equation}
L_{\text{shape cons}} = || \shape_i - \shape_j ||.
\end{equation}
Furthermore, we want to guarantee that we get a valid 3D shape, i.e., the estimated pose and shape parameters of the parametric model lie in the space of valid poses and shapes respectively. To enforce this, we use the adversarial prior of Kanazawa~\etal~\cite{kanazawa2018end}, which factorizes the model parameters into: (i) pose parameters $\pose$, (ii) shape parameters $\shape$, and (iii) per-part relative rotations $\pose_i$, that is one 3D rotation for each of the 23 joints of SMPL. In the end, we train a discriminator $D_k$ for each factor of the body model. The generator loss can then be expressed as:
\begin{equation}
L_{\text{adv prior}} = \sum_k (D_k(\vidt) - 1)^2.
\end{equation}
Depending on the availability of additional 2D keypoint annotations, we can also enforce that the 3D joints project close to the annotated 2D keypoints. We get the projection of the 3D joints $X$ to the 2D locations $x$, based on Eq.~\ref{eq:proj}. Then, the 2D-3D consistency can be expressed as:
\begin{equation}
L_{\text{2D}} =  ||{x} - x_{\mathrm{gt}}||,
\end{equation}
where ${x}_{\mathrm{gt}}$ are the ground truth 2D joints. Finally, adding smoothness on the pose parameters is also possible, but we avoid it, to keep our approach more generic and applicable even in settings
where the frames are not consecutive.

\textbf{Multiple views}: 
When we have access to multiple views $i$ and $j$ of a subject at the same time instance, then all the above losses remain relevant. The main additional constraint we need to enforce is that the pose of the person is the same across all viewpoints. This could be incorporated, by simply forcing all the pose parameters to have the same value. In contrast to that, we observed that a loss applied directly on the mesh vertices behaves much better (Figure~\ref{fig:mview}). This can be formulated as a simple per-vertex loss:
\begin{equation}
L_{\text{mesh cons}} =  ||M_i - M_j||.
\end{equation}
Remember that $M_i,M_j$ do not include the global orientation estimates $R_i,R_j$, so both meshes are in the canonical orientation, meaning that we can compare them directly. This loss effectively reflects the more generic case, where no knowledge of the camera extrinsics is available for the multi-view system. If extrinsic calibration is also known, then we simply need to apply the global pose estimates $R_i,R_j$, transform the second mesh to the coordinate system of the first mesh and then use the same per-vertex loss.

\subsection{Complete pipeline} \label{complete}
Our network is trained using batches of images. When we want to use a short sequence in training, or a few time-synchronized viewpoints, we include all the frames of interest in the batch. Typically, for monocular video, we include five consecutive frames, while for multi-view images, we use as many viewpoints are available at a specific time instance (typically four for Human3.6M). Conveniently, during testing, we can process each frame independently, without the need for video or multi-view input.

Depending on the setting, and making sure that we are compatible with prior work, we can also augment our batches with images that have stronger supervision (e.g., full 3D pose is known). Since the texture consistency assumption alone keeps the problem pretty underconstrained, similar to prior works (e.g.,~\cite{rhodin2018learning, rhodin2018unsupervised}), we found that it was useful to have stronger supervision in at least a few examples. For fair comparisons, in the empirical evaluation, we make sure that we use the same, or strictly less annotations than what prior work is using.

\subsection{Shortcomings} \label{shortcomings}
Although we empirically demonstrate the significant value of TexturePose (Section~\ref{experiments}), it is fair to also identify some of the shortcomings of our approach. For example the constant appearance assumption can easily be violated (e.g., due to illumination or viewpoint changes). Moreover, motion blur is common and can also decrease the level of ``clean'' pixels we can benefit from. Finally, our approach makes an assumption that no object occludes the person. Since we do not account for the potential occlusions, we can easily fill the texture map with the texture of the occluding object. Although occlusions are not very typical in most of the images for the datasets we use, this can be a source of potential error given a new video for training. The work of Ranjan~\etal~\cite{ranjan2018learning} addresses a similar problem in the context of Structure from Motion, and we believe that a similar approach should be applicable in our setting as well.

\section{Empirical evaluation}\label{experiments}
In this Section, we summarize the empirical evaluation of our approach. First, we provide more details about the datasets we employ for training and evaluation (Subsection~\ref{sec:datasets}), and then we present quantitative (Subsection~\ref{sec:quantitative}) and qualitative results (Subsection~\ref{sec:qualitative}).

\subsection{Datasets}\label{sec:datasets}
For the majority of our ablation studies, we used the Human3.6M dataset~\cite{ionescu2014human3}. Additionally, we used training data from the MPII 2D human pose dataset~\cite{andriluka20142d}, while LSP dataset~\cite{johnson2010clustered} was employed only to evaluate our approach. In the Sup.Mat. we present more extensive experiments leveraging the recently introduced VLOG-People and InstaVariety datasets~\cite{kanazawa2018dynamics} for training, as well as the 3DPW dataset~\cite{von2018recovering} for evaluation.

\noindent
\textbf{Human3.6M}:
It is an indoor benchmark for 3D human pose estimation. It includes multiple subjects performing daily actions like Eating, Smoking and Walking. It provides videos from four calibrated, time-synchronized cameras, making it easy to evaluate the different aspects of our approach both in the monocular and the multi-view setting. For training, we used subjects S1, S5, S6, S7 and S8, unless otherwise stated. Being consistent with prior work~\cite{kanazawa2018end}, the evaluation is done on subjects S9 and S11, considering Protocol 1~\cite{rhodin2018unsupervised,rhodin2018learning} and Protocol 2~\cite{kanazawa2018end}.

\noindent
\textbf{MPII}:
It is an in-the-wild dataset for 2D human pose estimation, providing only the 2D joint locations for each person. Previous works~\cite{kanazawa2018end,pavlakos2018learning} typically employ this dataset because of the large number of 2D keypoint annotations. One typically unexplored advantage of this dataset is the fact that it also provides the neighboring frames of the video that includes the annotated frame. We see this as a large pool of unlabeled data, that we can leverage for free, and we demonstrate their effectiveness in training our models. We call this set ``MPII video'' and consists of the annotated frames for each video, along with four more frames (two before, two after), which come with no labels.

\noindent
\textbf{LSP}:
It is also an in-the-wild dataset for 2D human pose estimation, but of much smaller scale compared to MPII. We employ LSP only for evaluation, where we make use of its test set. Particularly, given our shape prediction, we project it back to the image and we evaluate silhouette and part segmentation accuracy. For this evaluation, we use the segmentation labels provided by~\cite{lassner2017unite}.

\subsection{Quantitative evaluation}\label{sec:quantitative}

\textbf{Ablative studies}:
We start with Human3.6M where we initially treat all the images as frames of monocular sequences. One strong baseline, inspired from the ``unpaired'' setting of~\cite{kanazawa2018end}, assumes that the network has access to the 2D joints for each image, and an independent dataset of 3D poses, but no image with corresponding 3D ground truth is available. We train the network with 2D reprojection loss, while we also enforce an adversarial prior for pose and shape parameters such that the predicted poses/shapes are close to the poses/shapes in the dataset. As we can see, in Table~\ref{tab:h36m_only} (first row), this gives us decent performance. If we also apply our texture consistency loss, over the frames of the short clips, then we get significant improvement (second row). Finally, to put these results into context, we also train the same architecture providing full 3D pose and shape ground truth for each image (third row). As expected, this ideal version is performing better, but our texture consistency loss managed to close the gap between the weakly- and the fully-supervised setting.

\begin{table}
\centering
\small
\hspace{-3mm}
\tabcolsep=0.85mm
\begin{tabular}{@{}lcc@{}}
\toprule
& P1 & P2 \\
\midrule
2D keypoints + GAN prior & 93.0 & 79.1 \\
2D keypoints + GAN prior + Texture Consistency & 80.2 & 76.2 \\
\midrule
3D Ground Truth & 64.8 & 63.9 \\
\bottomrule
\end{tabular}
\vspace{3mm}
\caption{The effect of texture consistency for monocular input on Human3.6M (Protocols 1 \& 2). The numbers are mean reconstruction errors in mm. Using only 2D annotations on each frame and an adversarial pose/shape prior, we get reasonable performance. Simply providing more video frames instead of single frames {\em without any additional annotation}, we are able to get an important performance improvement, because of the texture consistency loss. As a lower limit, we present results when the ground truth 3D pose and shape parameters are available for each image during training. Although this last version uses explicitly stronger annotations, we are able to shrink the gap between the baselines that train with 2D and 3D annotations respectively.}
\vspace{-3mm}
\label{tab:h36m_only}
\end{table}

A similar experiment attempts to investigate the effect of leveraging texture consistency, but this time from in-the-wild videos. To this end, we use the frames of MPII video applying our texture consistency. The results for our experiments are presented in Table~\ref{tab:h36m_mpii}. The initial baseline (first row) is the same as in Table~\ref{tab:h36m_only}, and uses full 3D ground truth from Human3.6M for training. The next thing we want to investigate is whether adding purely unlabeled video can improve performance. So, for the next baseline (second row), we provide no labels for the in-the-wild frames, but enforce texture consistency. Interestingly, the model does get improved just by seeing more unlabeled data simply by enforcing texture consistency. Unfortunately, we observed that although the performance improves for Human3.6M, when we apply the same model to in-the-wild images, it achieves mediocre results qualitatively. We believe that at least a few labels should be necessary to make the model generalize better. To this end, we conduct two more experiments, adding annotations for one frame of the MPII video sequences. In the first experiment (third row), we add the annotation for the frame, but no texture consistency loss is enforced, while for the second one (fourth row), we both add the annotation for the frame, and we activate the texture consistency loss. As we can see, adding the unlabeled frames helps by default when combined with a texture consistency loss, and gives a solid performance improvement, making the model appropriate both for Human3.6M and for in-the-wild images, as we will present later.

\begin{table}
\centering
\small
\hspace{-3mm}
\tabcolsep=0.85mm
\begin{tabular}{@{}lccccc@{}}
\toprule
& & & P1 & P2 \\
\midrule
H36M & & & 64.8 & 63.9 \\
H36M & + MPII videos (+texture) & & 60.1 & 58.6 \\
\midrule
H36M & & + MPII 2D & 54.1 & 51.6 \\
H36M & + MPII videos (+texture) & + MPII 2D & 51.3 & 49.7 \\
\bottomrule
\end{tabular}
\vspace{3mm}
\caption{Evaluation on the Human3.6M dataset (Protocols 1 \& 2), indicating the effect of TexturePose, when we incorporate in-the-wild videos (MPII) in our training. Adding unlabeled video frames and enforcing texture consistency (row 2) improves Human3.6M evaluation, but the qualitative performance for in-the-wild images is mediocre. If we add a sparse set of 2D keypoint annotations (row 3), the performance can improve. However, the most interesting aspect is that by simply adding the sparse 2D keypoints labels, the {\em unlabeled} video frames {\em and} enforcing texture consistency (row 4), we can improve performance even more, meaning that extra
unlabeled data can always be helpful.}
\label{tab:h36m_mpii}
\end{table}

The same findings extend also to the case that we add more video data that only contain automatic \mbox{pseudo-annotations}~\cite{kanazawa2018dynamics,arnab2019exploiting}. We present our results using two recently published datasets for training, i.e., VLOG-People and InstaVariety~\cite{kanazawa2018dynamics} in the Sup.Mat., along with the 3DPW dataset~\cite{von2018recovering} for additional evaluation.

\textbf{Comparison with the state-of-the-art}:
For the comparison with the state-of-the-art, we use our best model from the previous experiment (last row of Table~\ref{tab:h36m_mpii}). The results are presented in Table~\ref{tab:h36m}. Our method outperform the previous baselines. Of course, we use MPII video for training which is not used by the other approaches, but making it possible to leverage the unlabeled frames is possible due to the texture consistency loss, which is one of our contributions. At the same time, other approaches also employ explicitly more annotations than we do, e.g.,~\cite{kanazawa2018end} has access to more images with 2D annotations (COCO~\cite{lin2014microsoft}) and 3D annotations (MPI-INF-3DHP~\cite{mehta2017monocular}), which we do not use.

\begin{table}
\centering
\small
\hspace{-3mm}
\tabcolsep=0.85mm
\begin{tabular}{@{}lc@{}}
\toprule
& Rec. Error \\
\midrule
Lassner~\etal~\cite{lassner2017unite} & 93.9 \\
Pavlakos~\etal~\cite{pavlakos2018learning} & 75.9 \\
NBF~\cite{omran2018neural} & 59.9 \\
HMR~\cite{kanazawa2018end} & 56.8 \\
Kanazawa~\etal~\cite{kanazawa2018dynamics} & 56.9 \\
Arnab~\etal~\cite{arnab2019exploiting} & 54.3 \\
Kolotouros~\etal~\cite{kolotouros2019convolutional} & 50.1 \\
Ours & \bf{49.7} \\
\bottomrule
\end{tabular}
\vspace{3mm}
\caption{Evaluation on the Human3.6M dataset (Protocol 2). The numbers are mean reconstruction errors in mm. We compare with regression approaches that output a mesh of the human body. Our approach achieves state-of-the-art results.}
\label{tab:h36m}
\end{table}

Moreover, we evaluate our approach on the LSP dataset using the same model (which has never been trained with images from LSP). Although in-the-wild, this dataset gives us access to segmentation annotations, so that we can evaluate shape estimation implicitly through mesh reprojection. The complete results are presented in Table~\ref{tab:lsp}. Here, we outperform the regression-based baseline of~\cite{kanazawa2018end} which is more relevant to ours and we are also very competitive to the optimization-based approaches, which explicitly optimize for the image-model alignment, so they tend to perform better under these metrics.

\begin{table}
\centering
\small
\hspace{-3mm}
\tabcolsep=2.95mm
\begin{tabular}{@{}lcccc@{}}
\toprule
& \multicolumn{2}{c}{FB Seg.} & \multicolumn{2}{c}{Part Seg.} \\
\cmidrule{2-5}
& acc. & f1 & acc. & f1 \\
\midrule
SMPLify \emph{oracle}~\cite{bogo2016keep} & 92.17 & 0.88 & 88.82 & 0.67 \\
SMPLify~\cite{bogo2016keep} & 91.89 & 0.88 & 87.71 & 0.64 \\
SMPLify on~\cite{pavlakos2018learning} & 92.17 & 0.88 & 88.24 & 0.64 \\
\midrule
HMR~\cite{kanazawa2018end} & 91.67 & 0.87 & 87.12 & 0.60 \\
Ours & 91.82 & 0.87 & 89.00 & 0.67 \\
\bottomrule
\end{tabular}
\vspace{3mm}
\caption{Evaluation on foreground-background and six-part segmentation on the LSP test set. The numbers are accuracies and f1 scores. Our approach outperforms the strong regression-based baseline of~\cite{kanazawa2018end} across the Table, and it is very competitive to the optimization baselines based on SMPLify (which typically have advantage for tasks involving image-model alignment like this). The numbers for the first two rows are taken from~\cite{lassner2017unite}.}
\label{tab:lsp}
\end{table}

Finally, we also compare with baselines trained with multiple-views. We follow the Protocol of Rhodin~\etal~\cite{rhodin2018unsupervised,rhodin2018learning}, training with full 3D supervision for S1 and employ the other training users without any further annotations, other than extrinsic calibration. The results are presented in Table~\ref{tab:h36m_multi}. We successfully outperform both baselines. It is interesting that both~\cite{rhodin2018unsupervised,rhodin2018learning} are non-parametric approaches, and we are still able to outperform them, considering that strong non-parametric baselines~\cite{martinez2017simple,pavlakos2018ordinal} typically perform better than parametric approaches~\cite{kanazawa2018end,pavlakos2018learning} (at least under the 3D joints metrics). We believe that this is exactly because we are able to leverage cues that are not an option for 3D skeleton baselines, e.g., they cannot map texture to a skeleton figure, as we can do with a mesh surface.

\begin{table}
\centering
\small
\hspace{-3mm}
\tabcolsep=0.85mm
\begin{tabular}{@{}lccc@{}}
\toprule
& MPJPE & NMPJPE & Rec. Error \\
\midrule
Rhodin~\etal~\cite{rhodin2018learning} & n/a & 153.3 & 128.6 \\
Rhodin~\etal~\cite{rhodin2018unsupervised} & 131.7 & 122.6 & 98.2 \\
Ours & {\bf 110.7} & {\bf 97.6} & {\bf 74.5} \\
\bottomrule
\end{tabular}
\vspace{3mm}
\caption{Evaluation on Human3.6M (Protocol 1) for methods trained on multiple views. The numbers are mm in various metrics. We follow the protocol of~\cite{rhodin2018learning,rhodin2018learning}, using full 3D ground truth for S1, and leveraging the other subjects as unlabeled data, where only the camera calibration is known.}
\label{tab:h36m_multi}
\end{table}

\subsection{Qualitative evaluation}\label{sec:qualitative}

\begin{figure*}[!t]
	\centering
	\begin{subfigure}[]{.15\textwidth}
		\includegraphics[width=\textwidth]{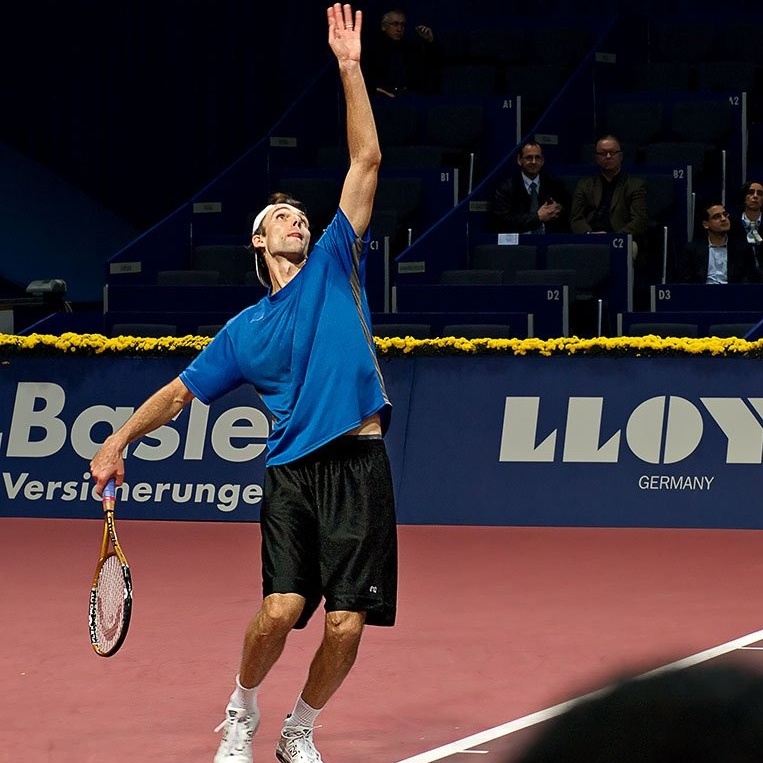}
	\end{subfigure}~
	\begin{subfigure}[]{.15\textwidth}
		\includegraphics[width=\textwidth]{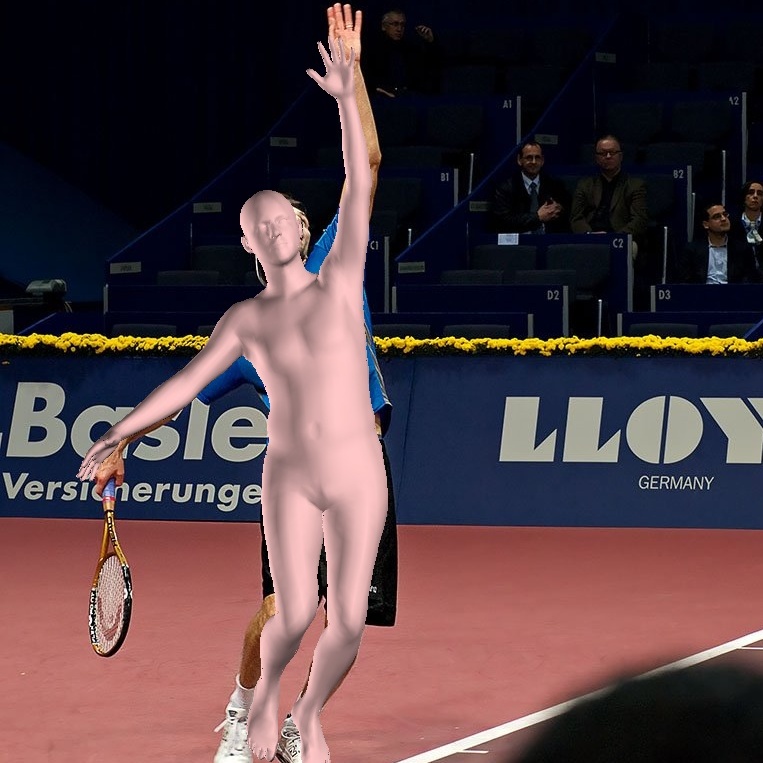}
	\end{subfigure}~
	\begin{subfigure}[]{.15\textwidth}
		\includegraphics[width=\textwidth]{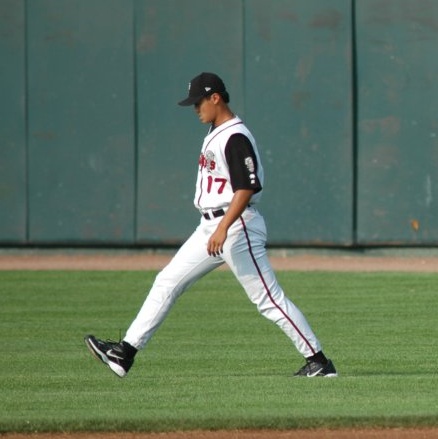}
	\end{subfigure}~
	\begin{subfigure}[]{.15\textwidth}
		\includegraphics[width=\textwidth]{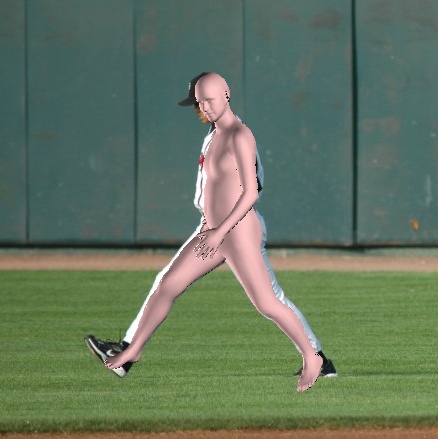}
	\end{subfigure}~
	\begin{subfigure}[]{.15\textwidth}
		\includegraphics[width=\textwidth]{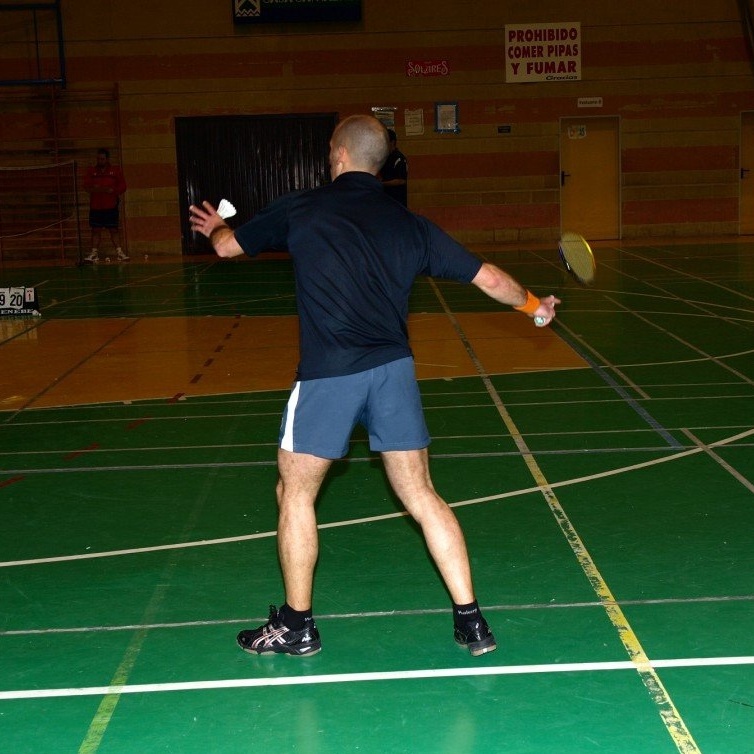}
	\end{subfigure}~
	\begin{subfigure}[]{.15\textwidth}
		\includegraphics[width=\textwidth]{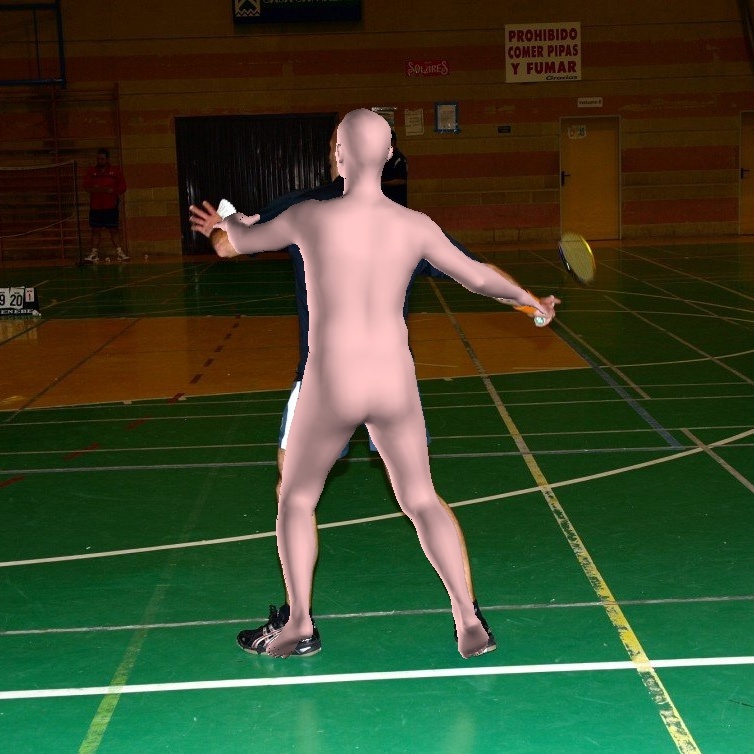}
	\end{subfigure}\\

	\begin{subfigure}[]{.15\textwidth}
		\includegraphics[width=\textwidth]{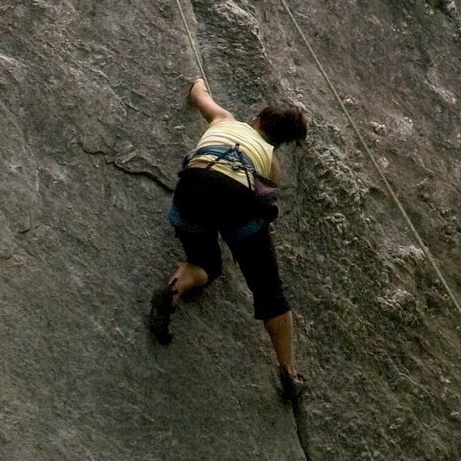}
	\end{subfigure}~
	\begin{subfigure}[]{.15\textwidth}
		\includegraphics[width=\textwidth]{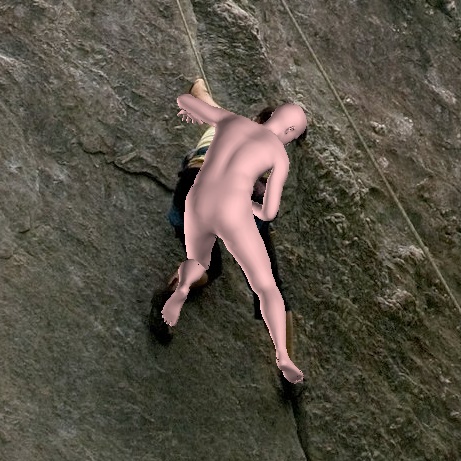}
	\end{subfigure}~
	\begin{subfigure}[]{.15\textwidth}
		\includegraphics[width=\textwidth]{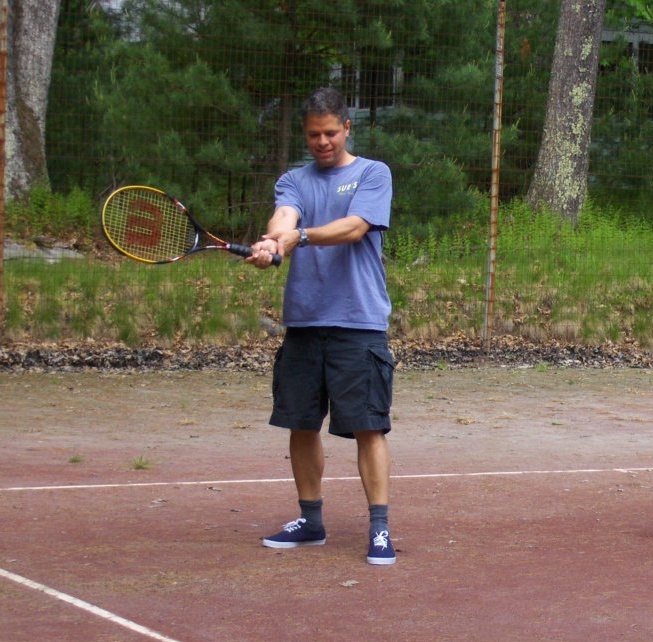}
	\end{subfigure}~
	\begin{subfigure}[]{.15\textwidth}
		\includegraphics[width=\textwidth]{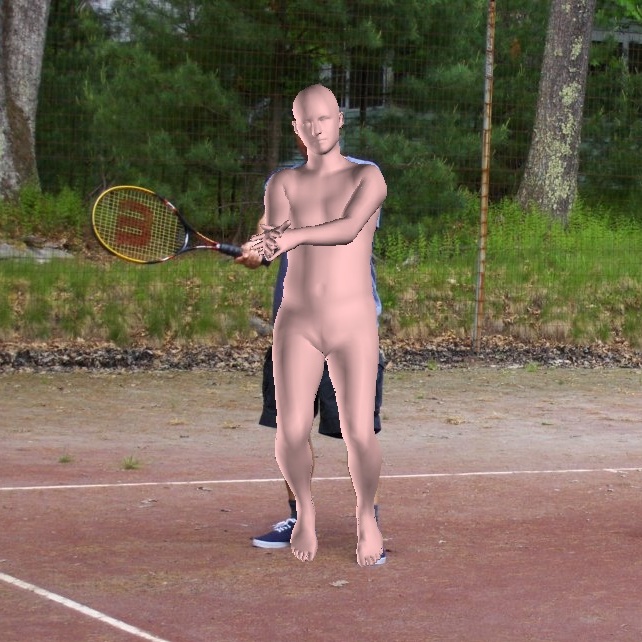}
	\end{subfigure}~
	\begin{subfigure}[]{.15\textwidth}
		\includegraphics[width=\textwidth]{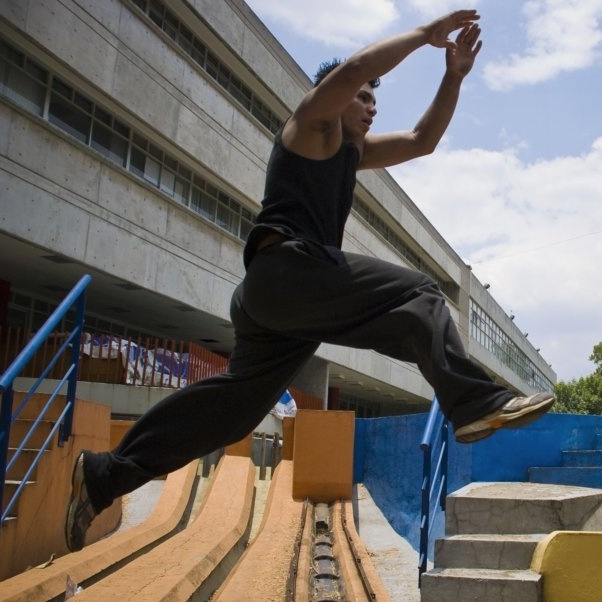}
	\end{subfigure}~
	\begin{subfigure}[]{.15\textwidth}
		\includegraphics[width=\textwidth]{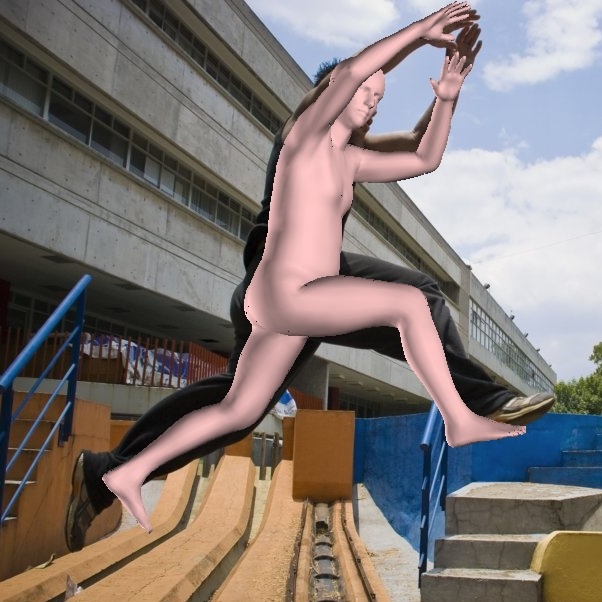}
	\end{subfigure}\\

	\begin{subfigure}[]{.15\textwidth}
		\includegraphics[width=\textwidth]{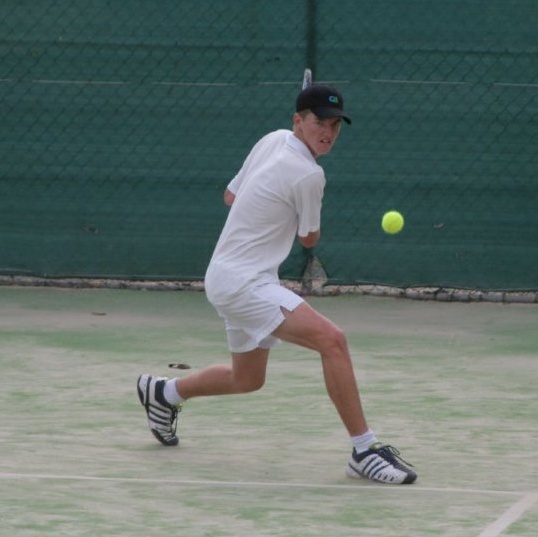}
	\end{subfigure}~
	\begin{subfigure}[]{.15\textwidth}
		\includegraphics[width=\textwidth]{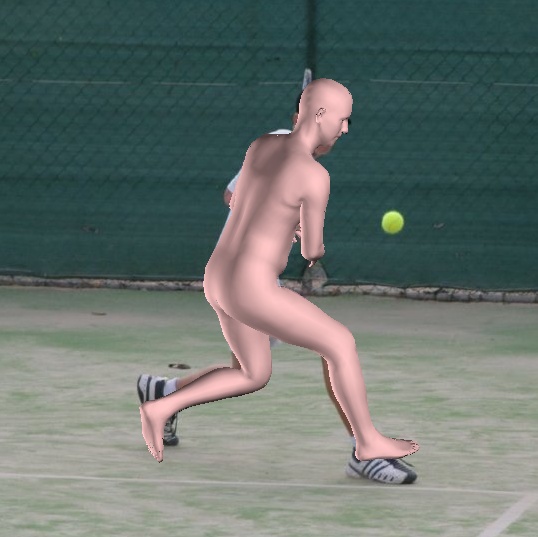}
	\end{subfigure}~
	\begin{subfigure}[]{.15\textwidth}
		\includegraphics[width=\textwidth]{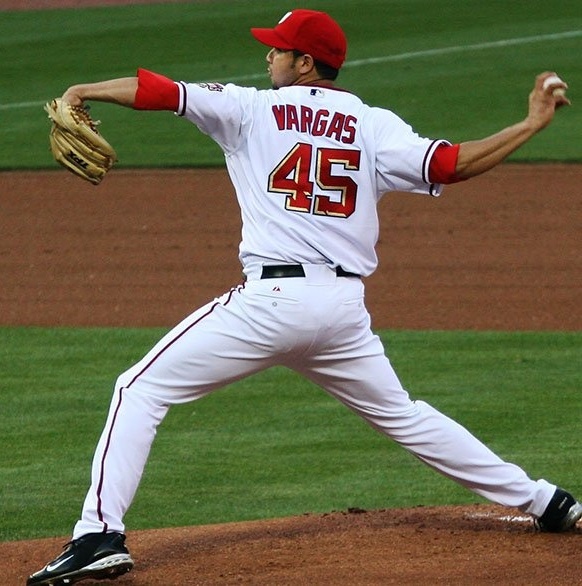}
	\end{subfigure}~
	\begin{subfigure}[]{.15\textwidth}
		\includegraphics[width=\textwidth]{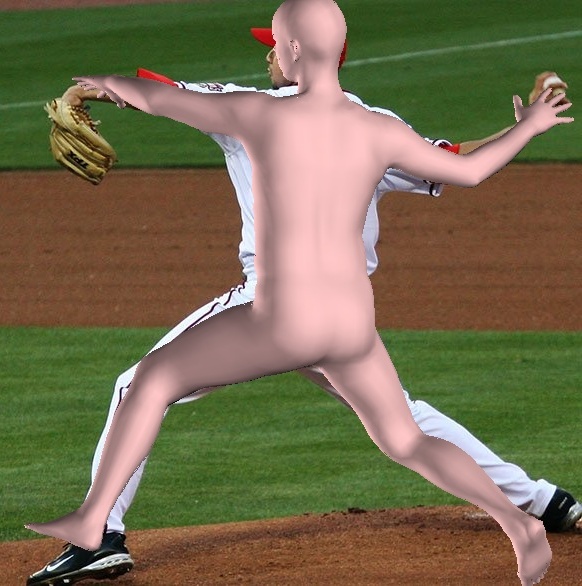}
	\end{subfigure}~
	\begin{subfigure}[]{.15\textwidth}
		\includegraphics[width=\textwidth]{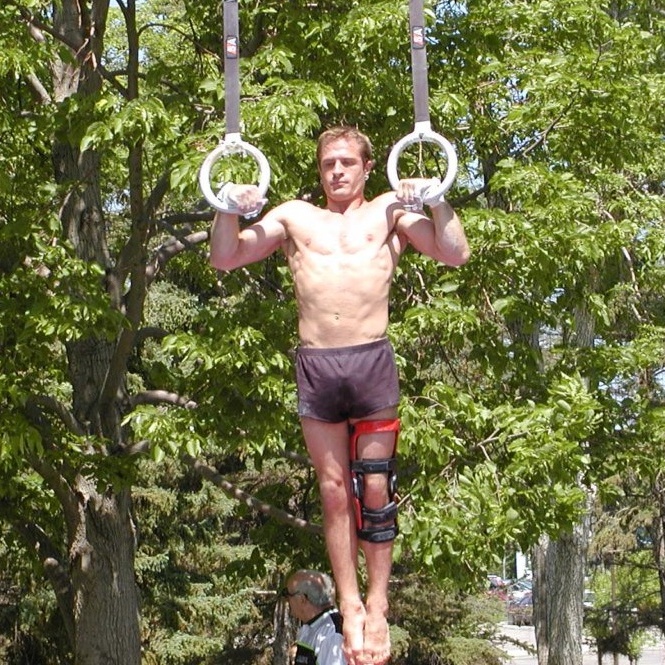}
	\end{subfigure}~
	\begin{subfigure}[]{.15\textwidth}
		\includegraphics[width=\textwidth]{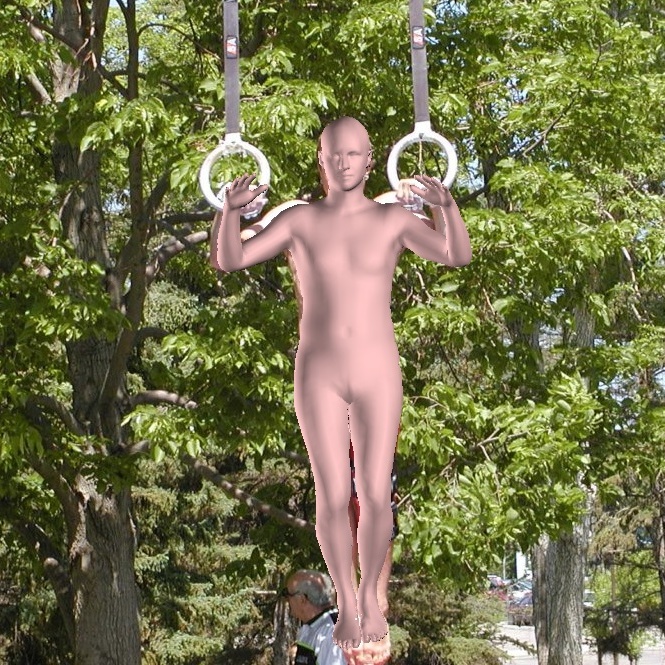}
	\end{subfigure}\\
	
	\begin{subfigure}[]{.15\textwidth}
		\includegraphics[width=\textwidth]{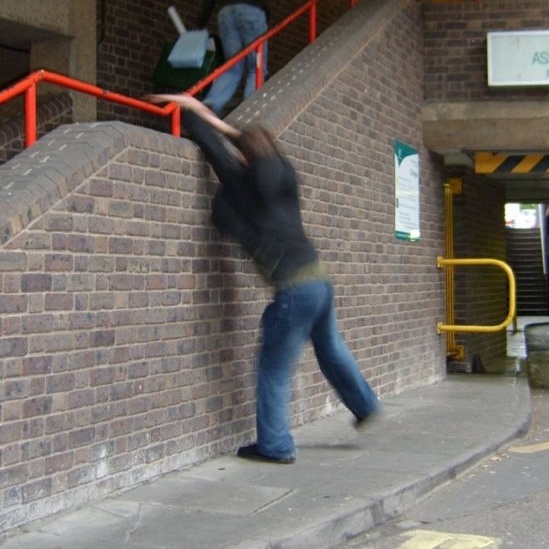}
	\end{subfigure}~
	\begin{subfigure}[]{.15\textwidth}
		\includegraphics[width=\textwidth]{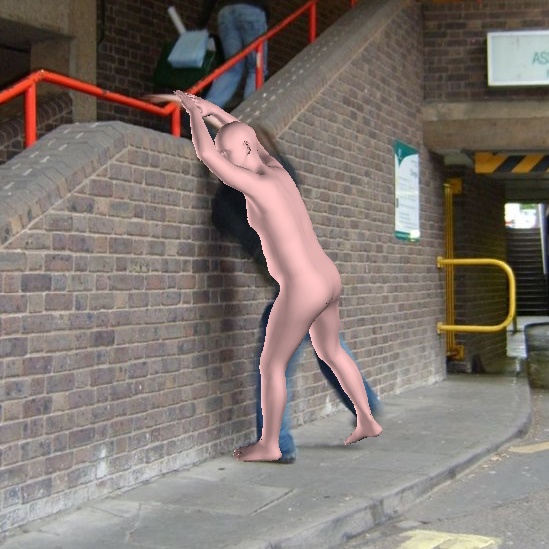}
	\end{subfigure}~
	\begin{subfigure}[]{.15\textwidth}
		\includegraphics[width=\textwidth]{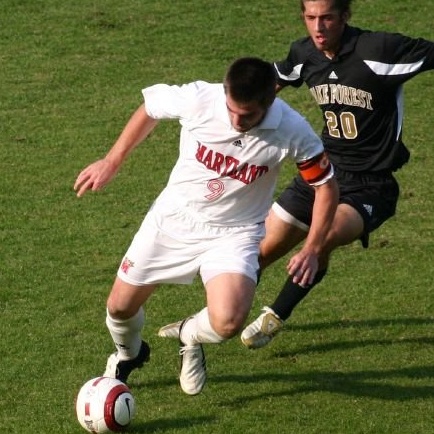}
	\end{subfigure}~
	\begin{subfigure}[]{.15\textwidth}
		\includegraphics[width=\textwidth]{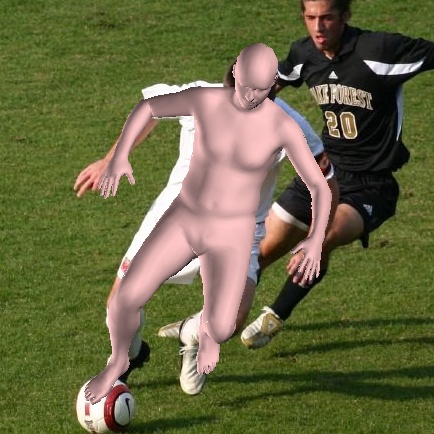}
	\end{subfigure}~
	\begin{subfigure}[]{.15\textwidth}
		\includegraphics[width=\textwidth]{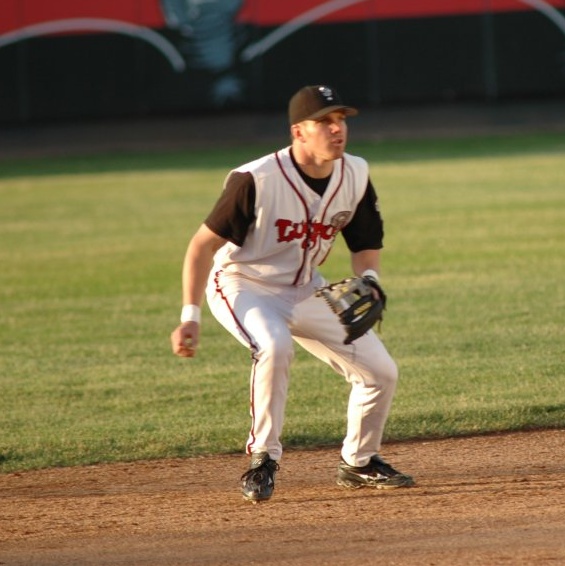}
	\end{subfigure}~
	\begin{subfigure}[]{.15\textwidth}
		\includegraphics[width=\textwidth]{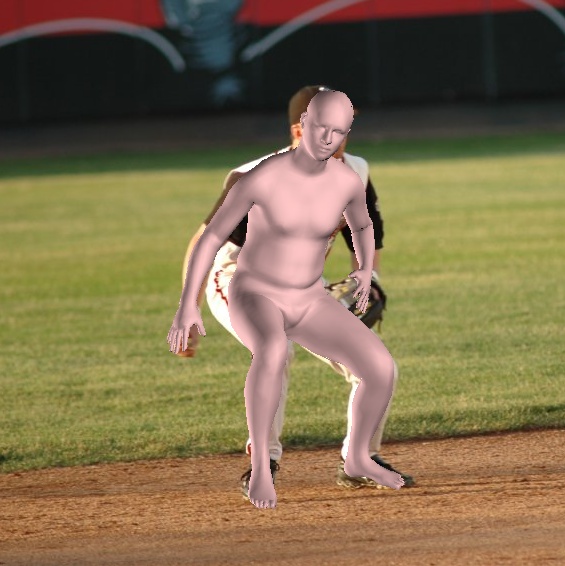}
	\end{subfigure}\\

	\begin{subfigure}[]{.15\textwidth}
		\includegraphics[width=\textwidth]{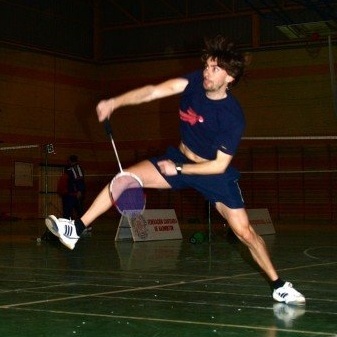}
	\end{subfigure}~
	\begin{subfigure}[]{.15\textwidth}
		\includegraphics[width=\textwidth]{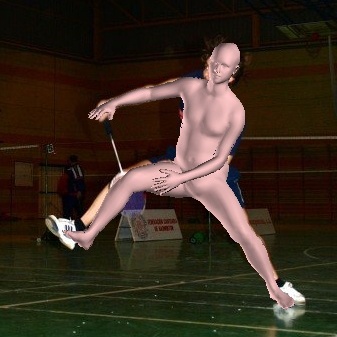}
	\end{subfigure}~
	\begin{subfigure}[]{.15\textwidth}
		\includegraphics[width=\textwidth]{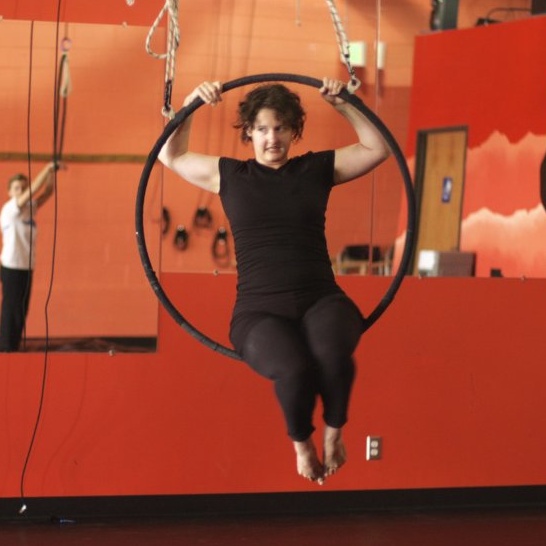}
	\end{subfigure}~
	\begin{subfigure}[]{.15\textwidth}
		\includegraphics[width=\textwidth]{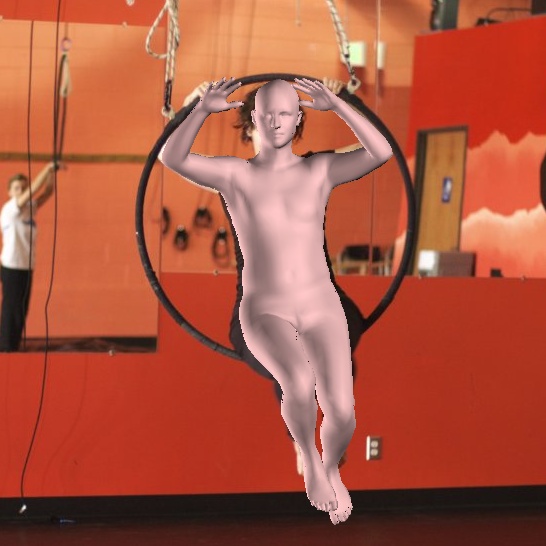}
	\end{subfigure}~
	\begin{subfigure}[]{.15\textwidth}
		\includegraphics[width=\textwidth]{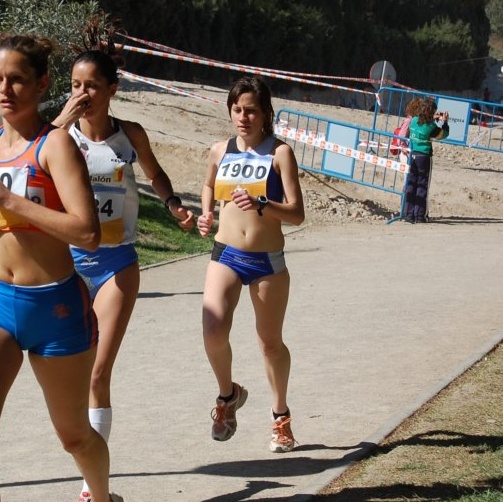}
	\end{subfigure}~
	\begin{subfigure}[]{.15\textwidth}
		\includegraphics[width=\textwidth]{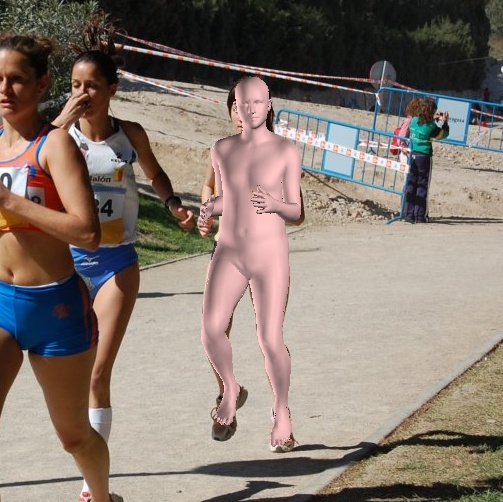}
	\end{subfigure}\\

	\begin{subfigure}[]{.15\textwidth}
		\includegraphics[width=\textwidth]{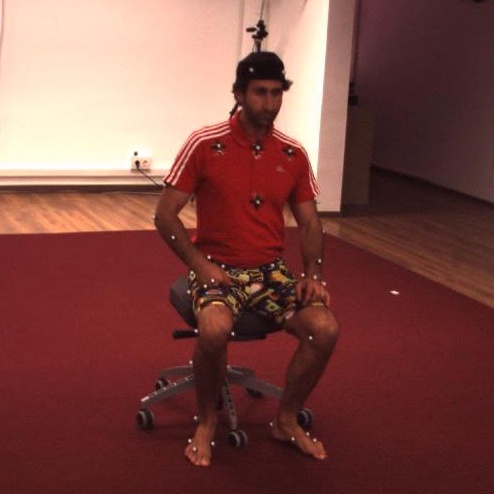}
	\end{subfigure}~
	\begin{subfigure}[]{.15\textwidth}
		\includegraphics[width=\textwidth]{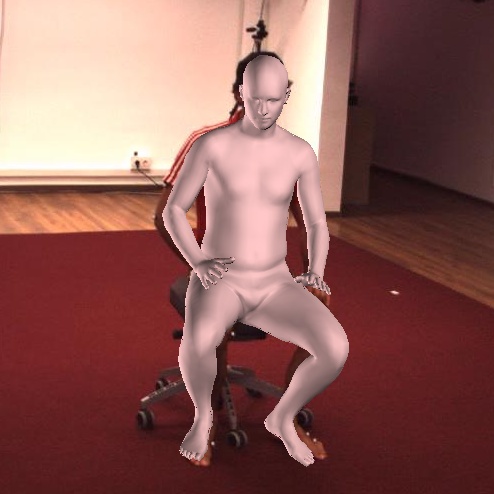}
	\end{subfigure}~
	\begin{subfigure}[]{.15\textwidth}
		\includegraphics[width=\textwidth]{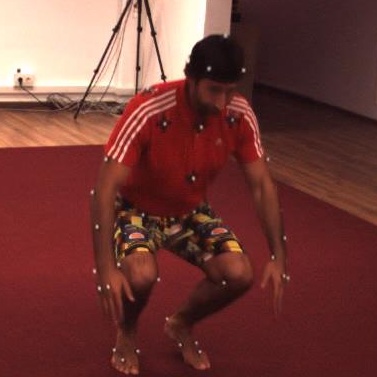}
	\end{subfigure}~
	\begin{subfigure}[]{.15\textwidth}
		\includegraphics[width=\textwidth]{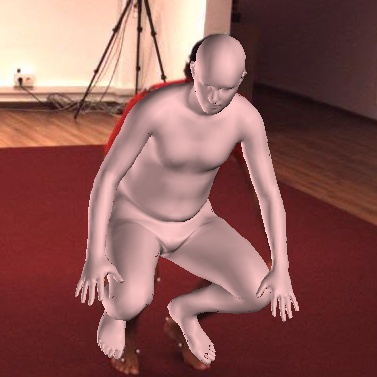}
	\end{subfigure}~
	\begin{subfigure}[]{.15\textwidth}
		\includegraphics[width=\textwidth]{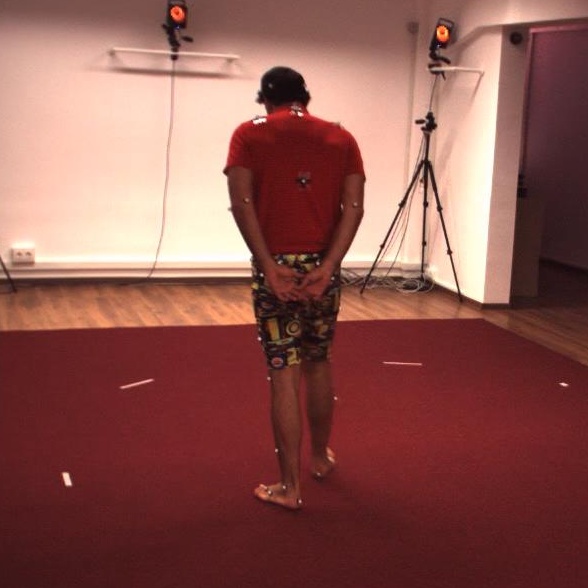}
	\end{subfigure}~
	\begin{subfigure}[]{.15\textwidth}
		\includegraphics[width=\textwidth]{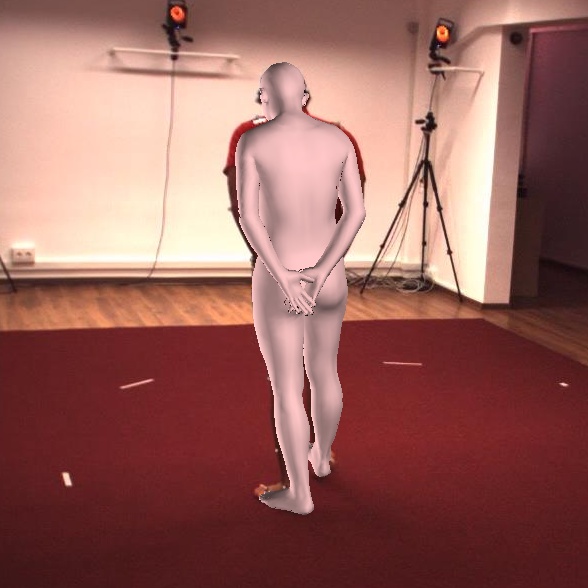}
	\end{subfigure}\\
	
	\begin{subfigure}[]{.15\textwidth}
		\includegraphics[width=\textwidth]{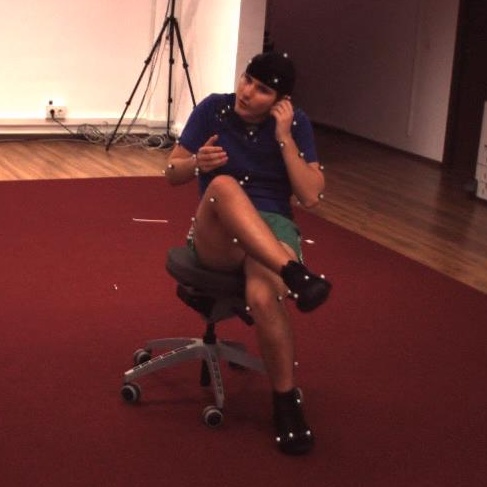}
	\end{subfigure}~
	\begin{subfigure}[]{.15\textwidth}
		\includegraphics[width=\textwidth]{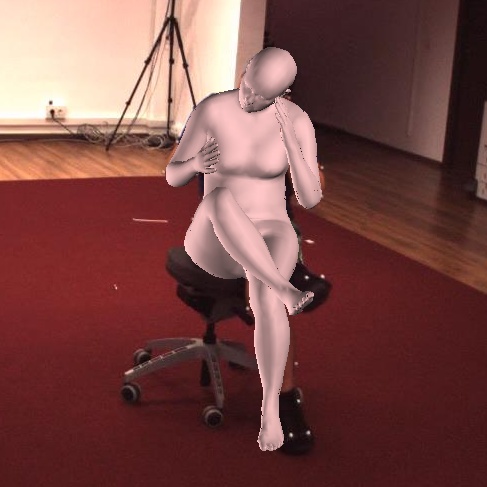}
	\end{subfigure}~
	\begin{subfigure}[]{.15\textwidth}
		\includegraphics[width=\textwidth]{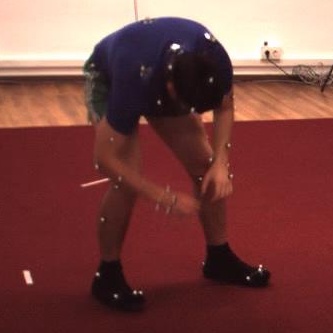}
	\end{subfigure}~
	\begin{subfigure}[]{.15\textwidth}
		\includegraphics[width=\textwidth]{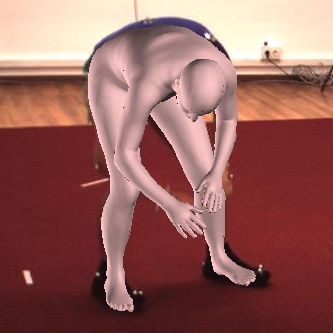}
	\end{subfigure}~
	\begin{subfigure}[]{.15\textwidth}
		\includegraphics[width=\textwidth]{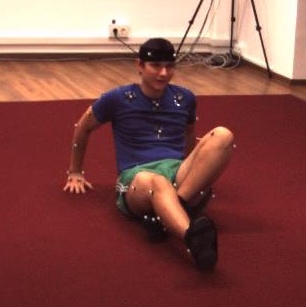}
	\end{subfigure}~
	\begin{subfigure}[]{.15\textwidth}
		\includegraphics[width=\textwidth]{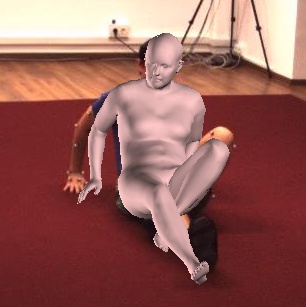}
	\end{subfigure}\\
	\caption{Qualitative Results. Rows 1-5: LSP dataset. Rows 6-7: H36M dataset}
\label{fig:results}
\end{figure*}

A variety of qualitative results of our results are provided in Figure~\ref{fig:results} as well as in the Sup.Mat.

\section{Summary}
In this paper, we presented TexturePose, an approach to train neural networks for model-based human pose estimation by leveraging supervision directly from natural images. Effectively, we capitalize on the observation that the appearance of a person does not change dramatically within a short video or for images from multiple views. This allows us to apply a texture consistency loss which acts as a form of auxiliary supervision. This generic formulation makes our approach particularly flexible and applicable in monocular video and multi-view images alike. We compare TexturePose with different baselines requiring the same (or larger) amount of annotations and we consistently outperform them, achieving state-of-the-art results across model-based pose estimation approaches. Going forwards, we believe that these weak supervision signals could really help us scale our training by leveraging weakly annotated or purely unlabeled data. Having already identified the shortcomings of our approach (Subsection~\ref{shortcomings}), it is a great challenge to identify ways to go beyond them.

\footnotesize
\noindent
{\bf Acknowledgements:} We gratefully appreciate support through the following grants: NSF-IIP-1439681 (I/UCRC), NSF-IIS-1703319, NSF MRI 1626008, ARL RCTA W911NF-10-2-0016, ONR N00014-17-1-2093, ARL DCIST CRA W911NF-17-2-0181, the DARPA-SRC C-BRIC, by Honda Research Institute and a Google Daydream Research Award.

{\small
\balance
\bibliographystyle{ieee_fullname}
\bibliography{egbib}
}

\end{document}